# Hierarchical Approach for Total Variation Digital Image Inpainting

[1]S.Padmavathi, [2]N.Archana, [3]K.P.Soman

[1] Department of Information Technology
[3] Head of CEN department
Amrita School of Engineering, Coimbatore, India.
1s_padmavathi@cb.amrita.edu, 2 archananallasamy@gmail.com ,
3kp_soman@amrita.edu

*Abstract*

*The art of recovering an image from damage in an undetectable form is known as inpainting. The manual work of inpainting is most often a very time consuming process. Due to digitalization of this technique, it is automatic and faster. In this paper, after the user selects the regions to be reconstructed, the algorithm automatically reconstruct the lost regions with the help of the information surrounding them. The existing methods perform very well when the region to be reconstructed is very small, but fails in proper reconstruction as the area increases. This paper describes a Hierarchical method by which the area to be inpainted is reduced in multiple levels and Total Variation(TV) method is used to inpaint in each level. This algorithm gives better performance when compared with other existing algorithms such as nearest neighbor interpolation, Inpainting through Blurring and Sobolev Inpainting.*

## I. INTRODUCTION

There are various real world situations where an image or photograph is damaged due to aging. The historical paintings in temples torn photographs of ancient history etc are few examples. The reconstruction of damaged images in a way that is undetectable for an observer who does not know the original image is known as inpainting. Inpainting can also be done digitally by using image-editing software. Virtually all inpainting operations require some kind of interaction with the user. Often the user needs to put down much effort in order to get a good result. There is a tool called cloning Brush to inpaint an image in Photoshop. The user specifies where the damage is and also specifies what has to be put in the damaged region. This process becomes more demanding when the damaged area is large in number and smaller in size. The objective of inpainting is to reconstitute the missing or damaged portions of the work without the users interaction.

Digital Image Inpainting performs inpainting digitally through image processing in some sense. It automates the process of filling and reduces the interaction with the user. However the user needs to specify where the damage is present in a given image. Ultimately, the only interaction required by the user is the selection of the region to be inpainted which is also called as mask. The user selected region is reconstructed by inpainting algorithm. In this paper a new algorithm for

International Journal of Computer Science, Engineering and Applications (IJCSEA) Vol.2, No.3, June 2012

automatic digital inpainting is proposed. This is based on the Hierarchical model that replicates the basic techniques used by professionals at each level of Hierarchy.

## II. STATE OF ART

In the inpainting literature the image to be inpainted is represented as shown in Fig.1. $u_0$ represents the image,     represents the region to be inpainted which is also called as target region and represents the boundary of    . The region ($u_0$-   ) is called as the source region where the pixel values are known and should not be altered by the inpainting algorithm. An algorithm using partial differential equations is presented in [1] is briefly discussed below. The basic idea is to iteratively fill    by prolonging the isophote lines arriving at    .

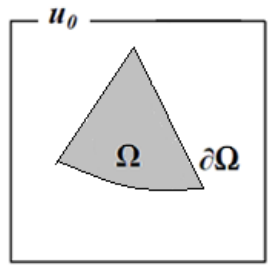

Figure 1: The image to be inpainted.

The difference however, lies in the goal of maintaining the angle of arrival. In order to maintain the angle of arrival, the direction of the largest spatial change is used. The direction may be obtained for each pixel along    by computing a discredited gradient vector and rotating this vector by    / 2 radians. Instead of using geodesic curves to connect the isophotes, the prolongation lines are progressively curved while preventing the lines from intersecting each other. This is done by using anisotropic diffusion. Each color channel of the image is treated as separate gray-level images. To estimate the variation of the intensity, a discrete two-dimensional laplacian is used. This estimation is spread in the isophote direction in order to maintain smooth intensity changes. In order to get a visually pleasing result it is important to propagate both the geometry (the structures in an image) and the photometry (the color values).

For removing *large* objects from digital photographs, exemplar based technique as discussed in [3] can be used. The result is an image in which the selected object has been replaced by a visually likely background that mimics the appearance of the source region. The technique used in [3] is capable of propagating both linear structure and two-dimensional texture into the target region with a single, simple algorithm. It uses a Patch based filling method which fills texture in a better way. Comparative experiments show that a simple selection of the fill order is necessary *and* sufficient to handle this task. This method performs equal to the previous techniques while restoring *small* scratches and it outperforms the earlier work while restoring *larger* objects, in terms of both perceptual quality and computational efficiency. The synthesis of regions does not produce reasonable results when similar patches do not exist and the algorithm is not designed to handle curved structures.

From [3] we adopt a unified scheme for determining the *fill order* of the target region. Pixels maintain a confidence value, which together with image isophotes, influence their fill priority.

## III.     METHODS USED IN HIERARCHICAL INPAINTING

### A. Interpolation Method

A naive way to recover the missing values is by performing an interpolation from the neighboring pixels. Each neighbor is at a unit distance from a pixel at (i,j). If in case (i,j) lies along the border of the image then some of its neighbors may lie outside the digital image also. To reconstruct the pixel the 8 neighbors, one in each direction as in the figure 2 are considered. Since the neighbors could belong to the missing area the procedure needs some modification. A simple nearest neighbor interpolation is adopted which considers neighbors of the source region only.

| (i-1,j-1) NW | (i-1,j) N | (i-1,j+1) NE |
|---|---|---|
| (i,j-1) W | (i,j) O | (i,j+1) E |
| (i+1,j-1) SW | (i+1,j) S | (i+1,j+1) SE |

Figure 2:   8 neighborhoods

### B. Inpainting by blurring

A simple way to recover the missing values of the image is to perform a blurring/smoothing. To progressively fill the hole, the blurring must be repeated. By considering the Gaussian filter for computation with symmetric/ periodic boundary conditions the blurring method is performed. So that the mask area is repeatedly blurred and the mask area is filled. This method does not reproduce the sharp edges.

### C. Inpainting by Sobolev Method [2]

A more mathematically sound way to perform inpainting is through regularization. The first method for improving generalization is called regularization. Inpainting problem could be thought of as minimizing the Total Variations in the inpainting area. Minimization problems are solved by computing Euler-Lagrange equation. In [2], the Sobolev gradient is used in the Euler Lagrange equation. They use a variational formulation and a gradient flow to converge to the solution. The inpainting region is considered as     and the inpainted region plus its boundary as,     ´. u(i,j) is considered to be a pixel in     ´ if u(k,l)     for some (k,l) with $0 \leq |i - k| + |j - l| \leq 2$. The pixels restricted to     and     ´ are denoted as $u_0$ and u´ as presented in [2]. The Sobolev gradient F(Du) is taken as in Eqn 3.1

$$F(Du) = D2u´ * D1 \quad u´ - D1u´ * D2 \quad u´ \qquad (3.1)$$

where $D1u_{i,j} = 0.5(u_{i+1,j} - u_{i-1,j})$,

$D2u_{i,j} = 0.5(u_{i,j+1} - u_{i,j-1})$

$u_{i,j} = 0.5(u_{i+1,j} + u_{i-1,j} + u_{i,j+1} + u_{i,j-1} - 4u_{i,j})$, for all (i,j) $\epsilon$   .

They discretized Euler-Lagrange equation as given in Eqn 3.2 ,

$$gEL = D1´v_{a1}(1) + D2´v_{a1}(2) + (D1´v_{a2}(1) + D2´v_{a2}(2)) \tag{3.2}$$

where Di´ denotes the adjoint of Di. And

$$v_{a1} = F(Du) * \begin{pmatrix} -D2\ u´ \\ D1\ u´ \end{pmatrix} \tag{3.3}$$

$$v_{a2} = F(Du) * \begin{pmatrix} -D2\ u´ \\ D1\ u´ \end{pmatrix} \tag{3.4}$$

By using the Sobolev gradient additional regularity is added to gradient. This added regularity means that, the image is changed by adding a scalar multiple of a smoother object, at each iteration.

## D. Hierarchical Inpainting using TotalVariation method (proposed method):

A simple way to recover the missing values of the image is to perform an interpolation. This method performs well in an uniform area, but results in blurring of the edges. The inpainting problem could be considered as a boundary value problem as given in Eqn. 3.5.

$$\left(\frac{\nabla u}{|\nabla u|}\right) = 0, \quad x \quad ; \quad u| = u^0| \tag{3.5}$$

As specified by[10] this could be considered as a Total variation problem. Let

$$v = (v^1, v^2) = \frac{'u}{|'u|} \tag{3.6}$$

Then the divergence is given by Eqn 3.7.

$$'v = \frac{\partial v^1}{\partial x} + \frac{\partial v^2}{\partial y} = v_e^1 - v_w^1 + v_n^2 - v_s^2 \tag{3.7}$$

Where e,w,n,s represents the directions East, West, North and South respectively. Let the $u_P$ represent the pixel in the direction 'P' with respect to the pixel O as shown in figure 2. Then, $v_e^1$ is given by the Eqn.3.8

$$v_e^1 = \frac{u_E - u_O}{|\ 'u_e|} \tag{3.8}$$

Where $|\ 'u_e| = \sqrt[2]{(u_E - u_O)^2 + [(u_{NE} + u_E - u_S - u_{SE})/4]^2}$ \tag{3.9}

Similar equations could be written for other directions. If $w_P = \frac{1}{|\nabla u_p|}$ then $u_O$ can be calculated as in Eqn. 3.10

$$u_O = {}_{!P}\left[\frac{\lambda_e(O).u_O^0}{\sum_P w_P + \lambda_e(O)}\right] \quad (3.10)$$

Where $\lambda_e(O)$ is the Lagrange multiplier. This TV method works well when the mask size is very small but fails as the mask size increases beyond 8 pixels.

The proposed method in this paper tries to keep the mask size smaller using a hierarchical approach and thereby utilizing the performance of the TV inpainting. Sampling the image at various resolutions reduces the image size along with the mask. The number of times the sampling is performed depends on the mask size under consideration. Once the mask size is reduced to the required level the TV inpainting is applied on the image. Hence the image at the lowest spatial resolution is inpainted first. The image at the next higher resolution is then considered for filling. The pixels that are inpainted at the lower resolution is updated in the current image which in turn maintains the mask size to a smaller value. This process is repeated until the actual sized image is inpainted. Since the unknown value has to be predicted from neighbors, the neighbors should be known. This forces the algorithm to fill the boundary pixels first. A count of the known pixels around it gives the confidence in filling a pixel. The confidence is high when count of known pixels is high. This confidence value is calculated for each pixels belonging to mask area. The interior pixels have a confidence value zero which specifies that they do not have known pixels around them. The pixels with higher confidence value are filled using equation 3.10. The value of Lagrange multiplier, $\lambda_e(O)$ is maintained such that the intensity from the known neighbors are used for inpainting. After filling the confidence value is updated and the process is repeated until all the pixels in the masked region are inpainted. Thus the mask is reconstructed / restored by TV method The algorithm given below describes the process of hierarchical TV inpainting..

1. Select the area to be inpainted, the mask
2. If mask size >Threshold T
    a. Increase the level count
    b. Down sample the image
    c. Store the image and mask; repeat step2
3. Inpaint the Image
    a. Find the boundary of the mask area
    b. Calculate the confidence value of all pixels on the boundary.
    c. Calculate the intensity using Eqn 3.10
    d. Update the boundary and repeat 3b and 3c
    e. Iterate until convergence threshold.
4. If not original image
    a. Copy the inpainted values from lower level
    b. Reduce the level count
    c. Update the mask values
    d. Go to step3
5. Display the image

## IV. EXPERIMENTAL RESULTS

Experiments have been conducted for various images with variable mask size and shape. The mask is chosen on uniform areas, high contrast areas etc. It is observed that for mask in uniform area interpolation method gives a quicker result with a quality equivalent to other methods. For multiple mask of relatively smaller area, Sobelev and TV inpainting methods produce better results when compared with interpolation methods. But for a single mask with larger area these methods do not perform well. The performance of inpainting algorithm is usually a subjective process. In order to show the improvement of the algorithm, the performance is made an objective process by replacing the user selected area in the input image with zero. The inpainting algorithm is then applied and the pixels in the mask area are reconstructed. The performance of inpainting algorithm is then measured by comparing the inpainted pixels and the original pixels of the image. Mean square Error as given in Eqn 4.1 and PSNR as given in Eqn. 4.2 are the metrics used for performance analysis.

$$MSE = \frac{1}{MN} \sum_{j=1}^{M} \sum_{k=1}^{N} (X_{j,k} - X'_{j,k})^2 \qquad 4.1$$

$$PSNR = 10 \log_{10} \frac{(2^n - 1)^n}{MSE} = 10 \log_{10} \left(\frac{255^2}{MSE}\right) \qquad 4.2$$

Where X and X' represent the original pixels and the inpainted pixels respectively.

The user selects the area to be inpainted from the input image. The mask is chosen near the edge. The methods as discussed in section III are used for inpainting. Figure 4.1 and 4.2 shows the inpainted results for two sample images. In each figure the images are ordered from (a) to (f) starting from top left and moving to right and then from bottom left to bottom right. The image on the top left (a) represents the original image. The image (b) on the right adjacent to it shows the image with the user selected area made zero. The result of applying nearest neighbor interpolation as discussed in section III A is shown in (c) which is the top rightmost image in the figure. The result of blur method as discussed in section III B is shown in (d) which is the bottom leftmost image in the figure. The result of Sobelev method as discussed in section III C is shown in (e) which is the bottom middle figure and the result of Hierarchical inpainting as discussed in section III D is shown in (f) which is the bottom rightmost image in the figure. A color image could be inpainted by considering the individual color channels separately. The result of inpainting a color image is shown in Figure 4.3. The image with mask area made zero is shown in Figure 4.3 (a). The result of hierarchical inpainting for three levels including the original sized image is shown in Figure 4.3(b), (c) and (d) respectively.

It could be observed that the nearest neighbor interpolation method simply propagates the pixels from the source region based on the confidence value of the boundary pixels. It also results in checkering effect. The replacement of intensity values are not well in the blur and Sobelev method. The Hierarchical method reproduces the intensity values in a better way. The algorithms are applied on images with increasing mask size for 110 images. The Mean square error and the PSNR are calculated for each category and the results are tabulated in Table 4.1 and 4.2 respectively. The corresponding graphs are plotted in Figure 4.4 and 4.5 respectively. From the graphs plotted it

International Journal of Computer Science, Engineering and Applications (IJCSEA) Vol.2, No.3, June 2012

could be seen that the Hierarchical method gives lesser error and lager PSNR than the rest of the methods.

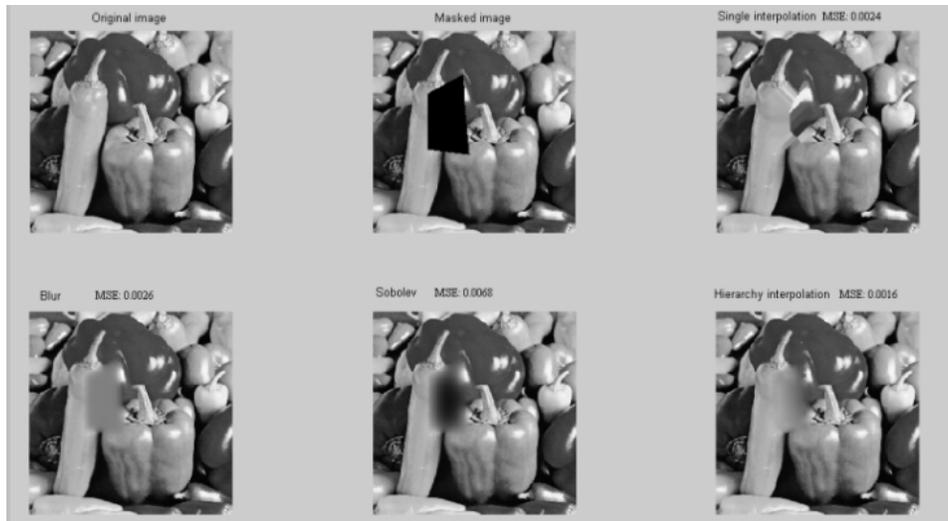

Figure 4.1 (a): Original image     (b): image with mask values zero    (c):image inpainted by interpolation
(d): image inpainted by blur    (e): image inpainted by Sobelev method  (f): image inpainted by Hierarchical method

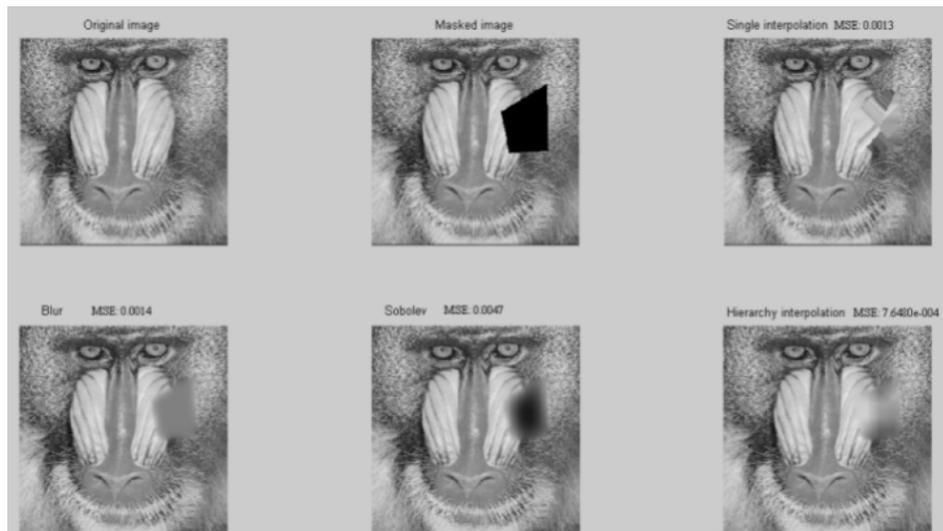

Figure 4.2 (a): Original image    (b): image with mask values zero   (c):image inpainted by interpolation
(d): image inpainted by blur   (e): image inpainted by Sobelev method   (f): image inpainted by Hierarchical method

Figure 4.1 and 4.2 - Sample output showing original image and inpainted images

179

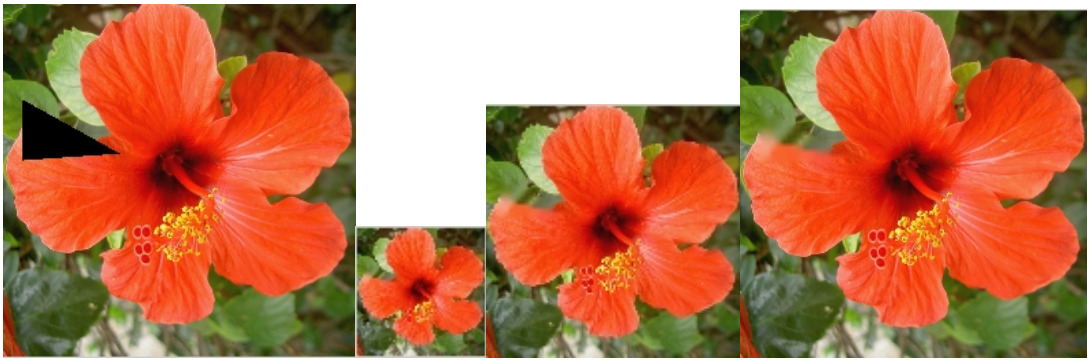

Figure 4.3 (a): color image with mask   (b)Inpainted at level 2   (c)Inpainted at level 1   (d)Inpainted at level 0, original size

Figure 4.3: Hierarchical inpainting of color image at various levels

TABLE 4.1: MSE FOR INPAINTING METHODS

| % of mask area reconstructed | Mean Square Error | | | |
|---|---|---|---|---|
| | *Neighbor Interpolation* | *Blur* | *Sobolev* | *Hierarchical Inpainting* |
| 2.1 | 8.36E-04 | 0.002 | 0.0047 | 4.10E-04 |
| 3 | 3.27E-04 | 6.41E-04 | 0.0039 | 2.65E-04 |
| 3.24 | 7.21E-04 | 0.0013 | 0.0011 | 7.07E-04 |
| 5.11 | 0.0016 | 0.0013 | 0.0042 | 9.03E-04 |
| 7.58 | 0.0026 | 0.0021 | 0.0093 | 0.0015 |
| 9.09 | 0.0099 | 0.0068 | 0.0087 | 0.0065 |
| 10.18 | 1.60E-03 | 0.0036 | 0.0051 | 0.0013 |
| 14.53 | 0.0065 | 0.0063 | 0.0084 | 0.005 |

TABLE 4.2: PSNR FOR INPAINTING METHODS

| % of mask area reconstructed | Peak Signal to Noise Ratio | | | |
|---|---|---|---|---|
| | *Neighbor Interpolation* | *Blur* | *Sobolev* | *Hierarchical Inpainting* |
| 2.1 | 78.91 | 75.1531 | 71.441 | 82.0033 |
| 3 | 82.9861 | 80.065 | 72.22 | 83.8968 |
| 3.24 | 79.5519 | 76.9995 | 77.5475 | 79.6338 |
| 5.11 | 76.2243 | 77.05 | 71.9142 | 78.5723 |
| 7.58 | 73.9514 | 74.936 | 68.4276 | 76.4308 |
| 9.09 | 68.1894 | 69.7901 | 68.7176 | 69.9691 |
| 10.18 | 76.0088 | 72.5556 | 71.0614 | 76.8772 |
| 14.53 | 69.9901 | 70.1694 | 68.8709 | 71.164 |

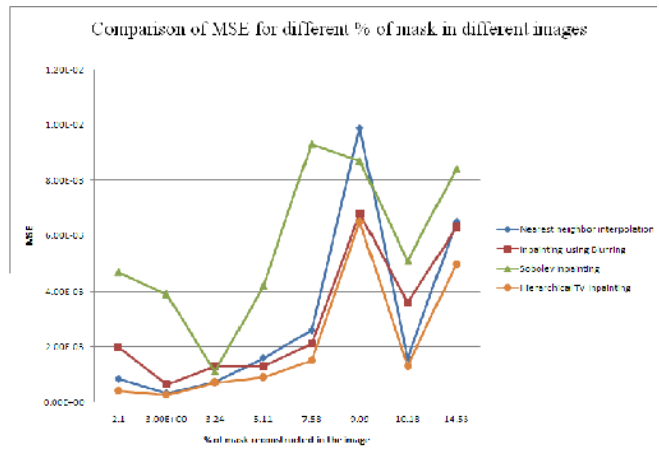

Figure 4.4 Plot of MSE for Inpainting methods

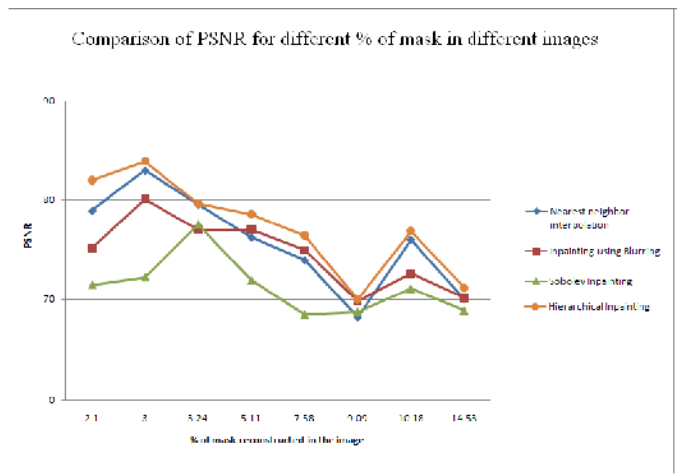

Figure 4.5 Plot of PSNR for Inpainting methods

The performance of the algorithm is not only influenced by the mask size but also it depends on the location of the region. If the regions are selected from high contrast areas then the error produced is also high. These are identified as peaks in the graph.

## V. CONCLUSION

Digital image inpainting offers a digital technique for restoring a damaged image. The algorithm requires the user to specify the damaged portion manually. It generates the damaged portions using other portions of the same image. It cannot generate a portion which is not available in the undamaged portions. Existing methods concentrate on images with smaller damaged portions. The quality of performance drops as the mask size increase. Hierarchical method proposed in this paper tries to utilize the advantage of the TV method while keeping the mask size less than a predefined value all the time. Though this method produces the gray levels better than other methods it results in some amount of blurring as the number of levels increases.

## VI. ACKNOWLEDGMENT

We would like to acknowledge, Jeevitha.K, Suganya.S and Priyalakshmi.B for their help in coding and generating the results.